\PassOptionsToPackage{table}{xcolor}
\documentclass[acmtog]{acmart}

\acmSubmissionID{624}

\usepackage[utf8]{inputenc} 
\usepackage[T1]{fontenc}    
\usepackage{url}            
\usepackage{amsfonts}       
\usepackage{nicefrac}       
\usepackage{microtype}      
\usepackage{xcolor}         
\usepackage{subcaption}
\usepackage[font=small]{caption}
\usepackage{float}
\usepackage{lscape}
\usepackage[lined,ruled,linesnumbered]{algorithm2e}
\usepackage{multirow}
\usepackage{enumitem}
\usepackage{bm}
\usepackage{mathtools}
\usepackage{color}
\usepackage{colortbl}
\usepackage{siunitx} 
\usepackage{graphicx}
\usepackage{listings}
\usepackage{xspace}

\renewcommand{\arraystretch}{1.3}

\def\ie{i.e.,\ }               

\definecolor{LavenderBlue}{rgb}{0.7020,    0.8039,    0.8902}

\newcommand{\norm}[1]{\left\lVert#1\right\rVert}

\usepackage[ruled]{algorithm2e} 

\SetAlFnt{\small}
\SetAlCapFnt{\small}
\SetAlCapNameFnt{\small}
\SetAlCapHSkip{0pt}

\author{Yun-Chun Chen}
\affiliation{
  \institution{University of Toronto}
  \country{Canada}
  \ and
  \institution{Adobe Research}
  \country{USA}
}
\orcid{0000-0003-0252-7503}
\email{ycchen@cs.toronto.edu}

\author{Selena Ling}
\affiliation{
  \institution{University of Toronto}
  \country{Canada}
  \ and
  \institution{Adobe Research}
  \country{USA}
}
\orcid{0000-0001-6458-4488}
\email{selena.ling@mail.utoronto.ca}

\author{Zhiqin Chen}
\affiliation{
  \institution{Adobe Research}
  \country{USA}
}
\orcid{0000-0001-7835-1618}
\email{zchen@adobe.com}

\author{Vladimir G. Kim}
\affiliation{
  \institution{Adobe Research}
  \country{USA}
}
\orcid{0000-0002-3996-6588}
\email{vokim@adobe.com}

\author{Matheus Gadelha}
\affiliation{
  \institution{Adobe Research}
  \country{USA}
}
\orcid{0000-0002-4971-7980}
\email{gadelha@adobe.com}

\author{Alec Jacobson}
\affiliation{
  \institution{University of Toronto and Adobe Research}
  \country{Canada}
}
\orcid{0000-0003-4603-7143}
\email{jacobson@cs.toronto.edu}

\copyrightyear{2024} 
\acmYear{2024} 
\setcopyright{acmlicensed}\acmConference[SA Conference Papers '24]{SIGGRAPH Asia 2024 Conference Papers}{December 3--6, 2024}{Tokyo, Japan}
\acmBooktitle{SIGGRAPH Asia 2024 Conference Papers (SA Conference Papers '24), December 3--6, 2024, Tokyo, Japan}
\acmDOI{10.1145/3680528.3687630}
\acmISBN{979-8-4007-1131-2/24/12}



\begin{document}

\title[Text-guided Controllable Mesh Refinement for Interactive 3D Modeling]{%
Text-guided Controllable Mesh Refinement for Interactive 3D Modeling
}

\begin{teaserfigure}
  \centering
  \includegraphics[width=\textwidth]{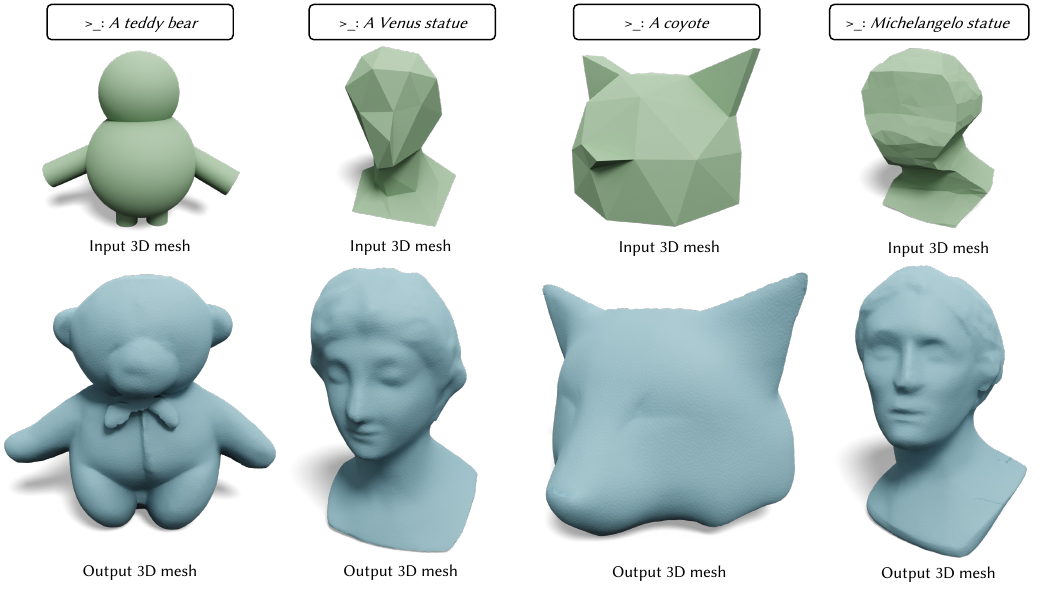}
  \vspace{-7.0mm}
  \caption{
  We present a method that adds geometric details to an input (coarse) 3D mesh through text guidance. 
  Our method can be applied to different types of input conditions. 
  From left to right, the input mesh is an assembly of six primitive shapes, a low-poly mesh, another low-poly mesh, and a mesh initialized by silhouette carving~\citep{space_carving}.
  }
  \label{fig:teaser}
\end{teaserfigure}

\begin{abstract}
We propose a novel technique for adding geometric details to an input coarse 3D mesh guided by a text prompt. 
Our method is composed of three stages. 
First, we generate a single-view RGB image conditioned on the input coarse geometry and the input text prompt. 
This single-view image generation step allows the user to pre-visualize the result and offers stronger conditioning for subsequent multi-view generation. 
Second, we use our novel multi-view normal generation architecture to jointly generate six different views of the normal images. 
The joint view generation reduces inconsistencies and leads to sharper details. 
Third, we optimize our mesh with respect to all views and generate a fine, detailed geometry as output. 
The resulting method produces an output within seconds and offers explicit user control over the coarse structure, pose, and desired details of the resulting 3D mesh. 
\end{abstract}

\maketitle

\section{Introduction}

We consider the problem of adding geometric details to 3D object meshes through text guidance. 
Text-to-3D generative modeling has been successfully applied to computer vision~\citep{DreamFusion,TECA}, computer graphics~\citep{CLIP_Mesh} and geometry processing~\citep{TextDeformer,FusionDeformer} applications. 
These methods focus on generating a 3D mesh directly from text~\citep{DreamFusion,ProlificDreamer} without providing users with the ability to control the coarse structure of the output shape, limiting their practical usefulness for artists who need careful control over the generation process. 
Other methods focus on generating textures for 3D meshes~\citep{TEXTure,TexFusion},
but they do not perform any changes in the geometry. 
While there are some approaches~\citep{TextDeformer,LatentNeRF} that provide control to users and are able to modify the geometry of a given shape, these methods are generally slow
since they rely on costly score distillation sampling~\citep{DreamFusion}.

\begin{figure}[t]
  \centering
  \includegraphics[width=\linewidth]{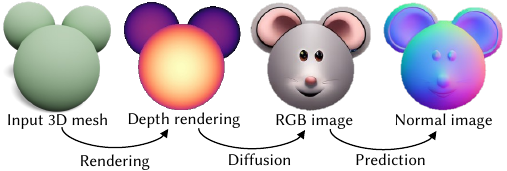}
  \vspace{-5.0mm}
  \caption{
  \textbf{Insight.}
  Using a depth/normal image as a control mechanism in large text-to-image models usually leads to images that contain more details than the ones present in the original shape. 
  Those details are so prominent that can even be captured by off-the-shelf shape estimation models.
  }
  \label{fig:insight}
\end{figure}

In this paper, we present a method that creates 3D object meshes with rich geometric details while allowing users to retain control over both the global shape structure (via an input coarse mesh) and the local geometric details (via an input text prompt).
Following the recent success of text-guided generative methods~\citep{TEXTure,DreamFusion,ProlificDreamer,LatentNeRF}, we build our method on large pre-trained text-to-image models~\citep{StableDiffusion} and use language to guide the generation of geometric details. 
Our formulation does not require paired coarse-geometry and fine-geometry training data but instead uses large pre-trained text-to-image models as supervision to guide the process of adding geometric details through a differentiable renderer. 
Our key insight stems from the fact that models trained to perform text-to-image generation
guided by depth information~\citep{ControlNet,T2I-Adapter} end up creating images
that contain additional geometric cues.
As we can see in Figure~\ref{fig:insight}, those cues are so prominent that even
off-the-shelf normal estimation models can extract them; \ie even though
the mouse image was generated from just three spheres, its normal estimation
(rightmost image) shows normals corresponding to a surface that depicts eyes, a nose and ears.
However, this process is only capable of creating details visible from a single
viewpoint whereas we want to add details to the whole visible surface of a given
shape.

Our method adds geometric details to an input mesh in three stages. 
The first stage generates a single-view RGB image based on the input text prompt and the input coarse mesh. 
This RGB image can be seen as a preview of how the geometric details that will be added to the input mesh look like. 
The second stage performs multi-view generation based on the single-view output from the first stage and the input coarse mesh. 
The third stage refines the geometric details of the input mesh based on the multi-view generation in the second stage. 
Since the output of each stage is non-abstract, human-understandable images or renderings, this property allows early moments before finish for users to decide to change parameters or go back and change/modify inputs. 
In addition, the first two stages only involve running inferences of pre-trained networks and the final stage operates directly on meshes. 
Each stage can be done within seconds, therefore allowing our method to be used to support interactive 3D modeling applications. 
As shown in Figure~\ref{fig:teaser}, our method can be used to add geometric details to objects assembled by primitive shapes, to low-poly meshes, and to meshes created from silhouette carving~\citep{space_carving}.

We compare our method with text-to-3D methods that allow for utilizing an input shape as guidance, i.e., Latent-NeRF~\citep{LatentNeRF}, Fantasia3D~\citep{Fantasia3D} and Magic3D~\citep{Magic3D}. 
Extensive experimental results show that our method produces shapes that have better geometric details and visual quality based on subjective human evaluations. 
Our results are also more consistent with input text conditions based on both subjective human evaluations and the objective CLIP~\citep{CLIP} similarity metric. 
At the same time, our method runs at least 90$\times$ faster than the competing methods. 

\section{Related Work}

Our work aims to generate shapes with rich geometric details from given coarse shapes while utilizing text prompts to guide the appearance of the generated details. 
Therefore, our work is closely related to methods that generate geometric details and text-to-3D generative methods. 
For text-to-3D methods, we focus on methods that utilize large pre-trained text-to-image diffusion models, which mainly include methods that adopt score distillation sampling (SDS) and methods based on multi-view image generation.

\begin{figure*}[t]
  \centering
  \includegraphics[width=\textwidth]{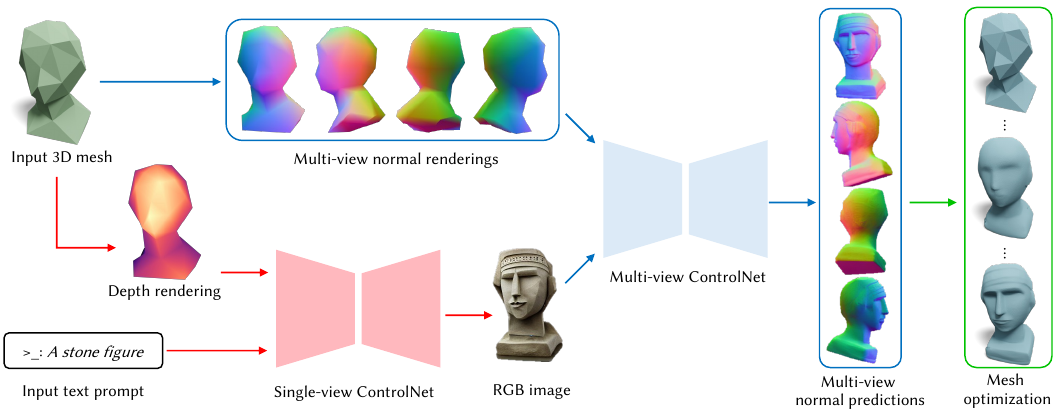}
  \vspace{-6.0mm}
  \caption{
  \textbf{Method overview.} 
  Our method consists of three stages: single-view generation, multi-view generation and mesh refinement/optimization. 
  Given an input mesh and an input text prompt, we first use a large-scale pre-trained diffusion model (highlighted in red) to generate an RGB image that respects the input conditions. 
  Next, we use a multi-view diffusion model (highlighted in blue) that takes as input the generated RGB image and the normal renderings of the input mesh and generates multi-view normals. 
  Finally, we use the generated multi-view normals to supervise the refinement of the input mesh. 
  }
  \label{fig:overview}
\end{figure*}

\paragraph{Geometric detail generation.}
With the polygon mesh representation, geometric details are typically generated on a coarse shape via transferring displacement maps or geometric texture patches from detailed shapes~\citep{Learning_detail_transfer,geometric_texture,Mesh_quilting}.
In addition, mesh subdivision powered by neural networks can transfer geometric details by learning to reconstruct local details from training examples~\citep{neural_subdiv,neural_prog_meshes}.
On the other hand, methods with voxel or other grid-structured representations are able to generate geometric details by replicating local patches from reference shapes~\citep{DECOR_GAN,DMTet,ShaDDR} via patch-wise generative adversarial networks~\citep{PatchGAN}.
These methods typically require detailed 3D shapes as geometric style references, which limits their capability when 3D shapes are scarce.
The progression on mesh-based differentiable rendering also enabled surface detail synthesis conditioned on the style of a reference image~\citep{Paparazzi} or input text~\citep{Text2Mesh,TextDeformer}.
However, these methods are slow due to per-shape optimization supervised by image diffusion models~\citep{Text2Mesh,TextDeformer}.

\paragraph{Text-to-3D with SDS}
The recent progress in text-to-image generative modeling has enabled a line of research works making use of large pre-trained text-to-image diffusion models to guide 3D generation through SDS, pioneered by DreamFusion~\citep{DreamFusion} that optimizes a neural radiance field (NeRF)~\citep{NeRF} for text-to-3D generation.
Several follow-up works explored optimizing NeRFs~\citep{LatentNeRF,Make-it-3d} and other representations with SDS, such as neural SDF~\citep{Textmesh}, meshes~\citep{Fantasia3D}, 3D Gaussian Splatting~\citep{DreamGaussian}, and hybrid representations~\citep{Magic3D,Magic123}.
Optimization-based methods re-train a model for each new text prompt, therefore suffering from a long inference time. Recent work addressed this issue by training a feed-forward model while using SDS to supervise the training~\citep{ATT3D}.

\paragraph{Text-to-3D with multi-view images.} 
Another line of works finetuned image diffusion models on large 3D object datasets, such as Objaverse~\citep{Objaverse}, to arrive at diffusion models capable of generating multiple images of the same 3D object viewed from different viewpoints. Such methods can produce multi-view images from a text prompt or a single-view image that could be generated by a text-to-image model. 
These multi-view images can be used to texture 3D meshes~\citep{TEXTure} or can be used to reconstruct a 3D shape via optimizing a 3D shape with differentiable rendering~\citep{zero123,zero123++,MVDream,Wonder3D,SyncDreamer} or via a feed-forward 3D shape reconstruction network~\citep{one2345,one2345++,Instant3D,LRM}.

Although prior text-to-3D methods are expressive in the final 3D generations, most of these methods lack any control mechanism besides the input text.
The unconstrained nature of these methods makes it difficult for 3D artists to incorporate them into their modeling process, which usually relies on fine-grained modifications.
Methods that allow a single-view image as an input could provide controllability to the 3D generative process to some extent. However, a single image is often ambiguous and lacks the ability to provide sufficient information to guide the generation of the desired structure.
Some optimization-based text-to-3D methods~\citep{LatentNeRF,Fantasia3D} allow having a coarse 3D shape to be the starting point of their generative procedure, therefore achieving generation with shape guidance. 
Yet they either produce poor geometry, as shown in Figure~\ref{fig:experiment} (b) and (d), or deviate from the structure of the input coarse shape, as in Figure~\ref{fig:experiment} (c). 
In addition, all methods take at least half an hour and up to several hours to generate an output due to their optimization-based nature. 
In contrast, we devise a multi-view ControlNet~\citep{ControlNet} conditioned on multi-view normal renderings of the coarse input shape, to provide global structure guidance.
Since our work does not rely on iterative optimization with image diffusion models, but instead relies on fast multi-view 3D reconstruction techniques~\citep{continuous_remeshing}, we are able to achieve significantly faster inference speed.

\section{Method}

Our method consists of three main stages: single-view RGB generation (Section~\ref{sec:single_view}), multi-view normal generation (Section~\ref{sec:multi_view}), and mesh refinement and optimization (Section~\ref{sec:refinement}).
Figure~\ref{fig:overview} shows the overview of our method.

Given a coarse triangle mesh $M_\text{coarse} = (V_\text{coarse}, F_\text{coarse})$ with vertex positions $V_\text{coarse}$ and faces $F_\text{coarse}$ and a text prompt $P$ as input, our goal is to use large pre-trained text-to-image models to generate images that have rich details and use them to refine the coarse mesh to create a triangle mesh $M_\text{fine} = (V_\text{fine}, F_\text{fine})$ that has additional geometric details as output.
Throughout this procedure, our method will rely on a rendering operator $\mathcal{R}$ that, given a mesh $M$ and a viewpoint $\theta$, 
yields a depth image $\mathcal{R}_d[M,\theta]$ and a normal image $\mathcal{R}_n[M, \theta]$.
We will also use the generative operators $\mathcal{G}$ that, given a control image $I$ and a text prompt $P$, output
a new image $\mathcal{G}[I, P]$ that follows both $I$'s structure and $P$'s description.
In practice, those operators correspond to deep generative models like ControlNet~\citep{ControlNet} and T2I-Adapters~\citep{T2I-Adapter}.

\subsection{Single-View RGB Generation}
\label{sec:single_view}

The goal of the first stage of our approach is to generate an RGB image that will act as a guide for the rest
of the mesh refinement process.
This is necessary for two reasons.
First, this image can be generated using large-scale text-to-image models that are general enough
to work with various prompts and types of objects while still being capable of creating detailed imagery.
Second, this initial procedure can act as a fast preview of the whole process -- the user can quickly
verify (in $\sim$3 seconds) if the image corresponds to the appearance they intend before continuing with the rest of the pipeline.
The process starts with the user providing a starting viewpoint $\theta_s$ (the view is given by the artist, i.e., the view that the object faces the camera), an initial mesh $M_\text{coarse}$ and the text prompt
$P$.
We use a standard depth-to-RGB ControlNet model $\mathcal{G}_d$ to create a color image $I_s$ that will be used in the rest
of the process:
\begin{equation}
    I_s = \mathcal{G}_d\big[\mathcal{R}_d[M_\text{coarse}, \theta_s], P\big].
\end{equation}

An important feature of using $\mathcal{G}_d$ to create $I_s$ is that its generative process happens through
a number of backward diffusion steps,
\ie iteratively denoising an image whose pixels are sampled from $\mathcal{N}(0,1)$.
Thus, we can allow the user to control how much $I_s$ deviates from $\mathcal{R}_d[M_\text{coarse}, \theta_s]$
by disabling the depth guidance in the later iterations of the denoising procedure. 
We refer readers to \citet{ControlNet} for further details about the model architecture and how the amount
of guidance can be set by the user.
In Section~\ref{sec:controlability} we discuss the effects of applying different levels of guidance throughout
the single-view and multi-view generation stages.

\begin{figure}[t]
  \centering
  \includegraphics[width=\linewidth]{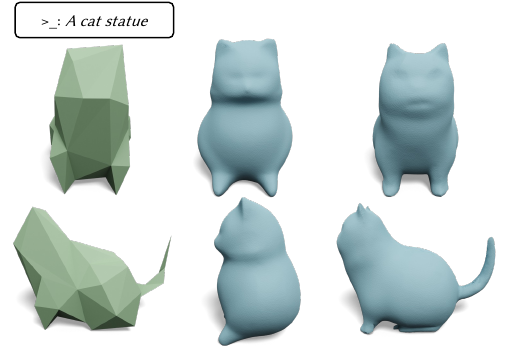} \\
  \vspace{-5.0mm}
  \begin{subfigure}[t]{0.325\linewidth}
    \caption*{Input 3D mesh}
  \end{subfigure}
  \hfill
  \begin{subfigure}[t]{0.325\linewidth}
    \caption*{Wonder3D}
  \end{subfigure}
  \hfill
  \begin{subfigure}[t]{0.325\linewidth}
    \caption*{Ours}
  \end{subfigure}
  \\
  \vspace{-3.0mm}
  \caption{
  \textbf{Multi-view control.}
  Our method is capable of generating details using the full initial shape as guidance.
  Notice how the back legs of the cat and its tail follow the input coarse mesh (in green).
  On the other hand, Wonder3D results yield reasonable renditions when visualized from the
  initial viewpoint $\theta_s$ (top row) but clearly fail to follow the coarse geometric guidance when
  seen from other views (bottom row).
  }
  \label{fig:exp-mv-control}
\end{figure}

After this process is complete, we have an RGB image $I_s$ that roughly follows the structure given by $M_\text{coarse}$ when
seen from $\theta_s$ while depicting the description $P$. 
Unfortunately, $I_s$ is generated based on a single-view depth rendering of the input mesh and has no information about what the mesh
we are trying to refine looks like from viewpoints other than $\theta_s$. 
One could use other methods that generate multi-view consistent images from a single input image (e.g., MVDream~\citep{MVDream} and Wonder3D~\citep{Wonder3D}), but they would not follow $M_\text{coarse}$ -- only what it looks
like when seen from $\theta_s$.
Examples of this behavior are depicted in Figure~\ref{fig:exp-mv-control} and Figure~\ref{app-fig:mv_control}.
While Wonder3D and other image-to-3D methods are capable of generating reasonable shapes, they fail to adhere
to any shape guidance not depicted in the input image.
We address this issue by designing a multi-view ControlNet architecture detailed in the following subsection. 

\subsection{Multi-view Normal Generation}
\label{sec:multi_view}

Until this point, we only used guidance from the initial viewpoint $\theta_s$. 
Now, we define a set of viewpoints 
$\Theta=\{ R_y(\alpha) | \alpha \in \{ 0, \frac{\pi}{4}, \frac{\pi}{2}, \pi, \frac{3\pi}{2}, \frac{7\pi}{4} \} \}$
where $R_y(\alpha)$ corresponds to
the viewpoint obtained by rotating the object $\alpha$ degrees along the upward axis from the starting viewpoint $\theta_s$. 
We propose a multi-view ControlNet $\mathcal{G}^\text{mv}$ that learns to generate multi-view consistent normal images $\{I_{n,\theta}\}_{\theta \in \Theta}$ 
conditioned on the single-view RGB image $I_s$ and on $|\Theta|$ multi-view normal renderings
$\{\mathcal{R}_n[M_\text{coarse}, \theta]\}_{\theta \in \Theta}$ of the input mesh $M_\text{coarse}$. 
More precisely,  we have
\begin{equation}
I_{n,\theta} = \mathcal{G}^\text{mv}_{n,\theta} \big[ \{\mathcal{R}_n[M_\text{coarse}, \theta]\}_{\theta \in \Theta}, I_s \big],
\end{equation}
where $\mathcal{G}^\text{mv}_{n,\theta}$ yields normal images of the
object seen from $\theta$.
Thanks to the additional information present in $I_s$ and the way $\mathcal{G}^{\text{mv}}$ is trained,
the normal images $I_{n,\theta}$ generated by our multi-view ControlNet preserve the coarse structure of $M_\text{coarse}$ while
containing details following the description $P$.
We build our $\mathcal{G}^\text{mv}$ on top of a Wonder3D~\citep{Wonder3D} network $\mathcal{W}^\text{mv}$ that is trained to generate multi-view consistent RGB and normal images given a single-view RGB image. 
We take inspiration from ControlNet~\citep{ControlNet} and design a control model $\mathcal{C}^\text{mv}$ that can be plugged into $\mathcal{W}^\text{mv}$ to allow
multi-view control in the existent multi-view generation network.

\begin{figure}[t]
  \centering
  \includegraphics[width=\columnwidth]{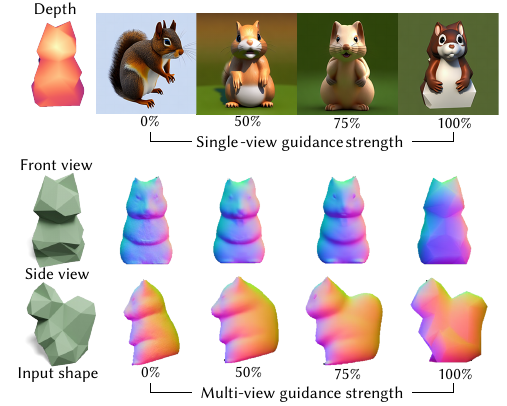}
  \vspace{-6.0mm}
  \caption{
  \textbf{Influence of guidance strength.} 
  The user can control how much detail can be generated by the system when setting the number of
  backward diffusion steps where the guidance will be used.
  This control is available during the single-view (top) and multi-view (bottom) stages.
  }
  \label{fig:guidance_strength}
\end{figure}

\begin{figure*}[t]
  \centering
  \includegraphics[width=\linewidth]{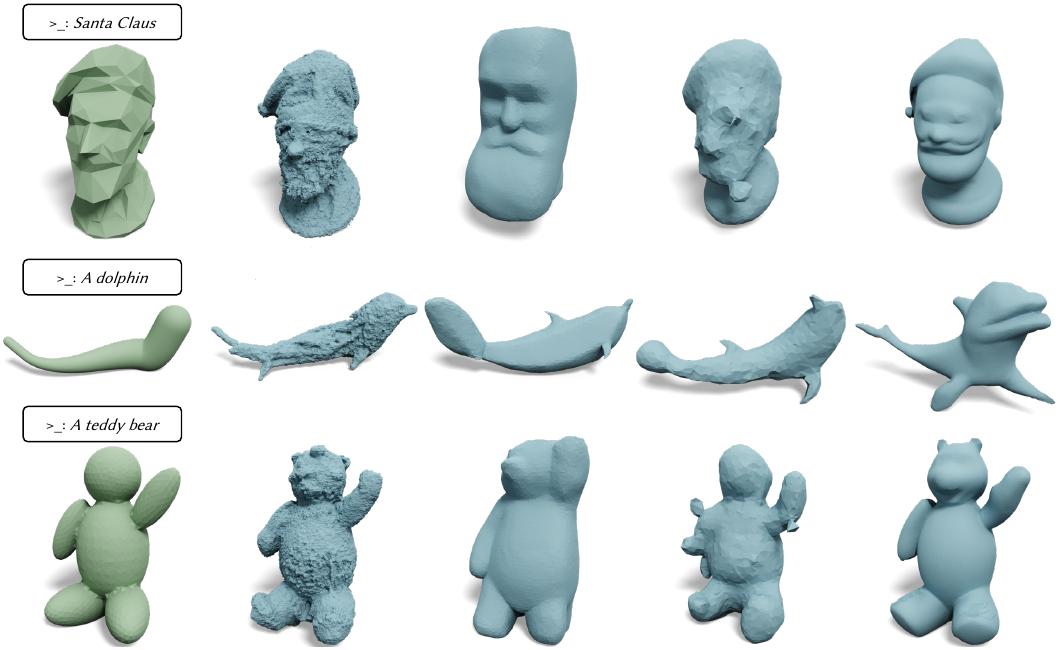} \\
  \vspace{-3.0mm}
  \begin{subfigure}[t]{0.196\linewidth}
    \caption{Input 3D mesh}
  \end{subfigure}
  \hfill
  \begin{subfigure}[t]{0.196\linewidth}
    \label{exp:sota-latentnerf}
    \caption{Latent-NeRF}
  \end{subfigure}
  \hfill
  \begin{subfigure}[t]{0.196\linewidth}
    \label{exp:sota-fantasia}
    \caption{Fantasia3D}
  \end{subfigure}
  \hfill
  \begin{subfigure}[t]{0.196\linewidth}
    \label{exp:sota-magic3d}
    \caption{Magic3D}
  \end{subfigure}
  \hfill
  \begin{subfigure}[t]{0.196\linewidth}
    \caption{Ours}
  \end{subfigure}
  \\
  \vspace{-3.0mm}
  \caption{
  \textbf{Qualitative results.} 
  Our method generates 3D meshes that have better geometric details and visual quality compared to state-of-the-art methods.
  }
  \label{fig:experiment}
\end{figure*}

\paragraph{Network details.} 
$\mathcal{W}^\text{mv}$ builds on pre-trained 2D Stable Diffusion~\citep{StableDiffusion} and enables multi-view RGB and normal generation from a single-view RGB image. 
The network takes as input a single-view RGB image and a camera pose, and outputs an RGB image and a normal image of the object depicted in the input image from the given camera pose. 
We refer readers to \citet{Wonder3D} for details about Wonder3D.
Our model $\mathcal{G}^\text{mv}$ consists of three components:
the original (frozen)  $\mathcal{W}^\text{mv}$,
a trainable $\mathcal{C}^\text{mv}$ copy of $\mathcal{W}^\text{mv}$,
and a normal encoder, which takes as input the set $\{\mathcal{R}_n[M_\text{coarse}, \theta]\}_{\theta \in \Theta}$. 
Specifically, $\mathcal{C}^\text{mv}$'s encoder takes as input the output of the normal encoder and outputs features that are then integrated into  $\mathcal{W}^\text{mv}$'s decoder's skip connections using zero-initialized convolutions. 
We follow the same training procedure as Wonder3D to train the normal encoder and $\mathcal{C}^\text{mv}$.

\paragraph{Training data.} 
To train our multi-view ControlNet, we generate training data using the LVIS subset of the Objaverse~\citep{Objaverse} dataset, which consists of around 32,000 objects. 
For each object, we use Blenderproc~\citep{Blenderproc} to render RGB images and normal images from 6 views, i.e., the front, left, right, back, front-left and front-right views. 
To encourage our multi-view ControlNet to learn to generate more details in the multi-view normal predictions, for every input normal
image we apply a Gaussian filter with a kernel size of 7. 
The goal of this blurring step is to ensure that the model is only aware of the coarse structure of the shape and not its specific details.
To train our multi-view ControlNet, we use the blurred normal image as input to the normal encoder and train the network to 
predict the unblurred (clear) normal image. 
This mechanism allows the model to learn to generate additional geometric details while following the coarse structure
of the blurred normals. 
This blurring operation is also applied at inference time. 

\paragraph{Implementation details.} 
We implement our network in PyTorch. 
The normal encoder consists of 5 convolutional layers. 
The first four convolution layers are randomly initialized, each of which is followed by a SiLU~\cite{SiLU} activation. 
The last convolution layer is zero-initialized.
The image resolution used to train the network is 256 $\times$ 256. 
We use a batch size of 192 and train the network for 20,000 steps. 
We use 4 NVIDIA A100 GPUs to train the network. 
The network training takes around 4 days. 

\subsection{Mesh Refinement and Optimization}
\label{sec:refinement}

Given the multi-view images generated by our multi-view ControlNet, there are multiple ways to reconstruct a 3D mesh from them. 
For instance, one can use multi-view images to fit a NeRF~\citep{NeRF,InstantNGP} or a neural SDF~\citep{NeuS}. 
However, computing these representations through gradient-based optimization takes a lot of time and, after the procedure is complete,
one would still need to convert them into meshes.
For our application, we cast mesh refinement directly as a mesh optimization problem. 
To make the mesh optimization pipeline fully differentiable, we use a differentiable rasterizer, nvdiffrast~\citep{nvdiffrast},
that allows us to quickly optimize the coarse mesh $M_{\text{coarse}}$ to follow the set of normal images $\{I_{n,\theta}\}_{\theta \in \Theta}$.

We will optimize the mesh $M$ initialized as $M = M_\text{coarse}$ and refine its geometry based on $\{I_{n,\theta}\}$ through
gradient descent. 
For each iteration, we optimize the mesh vertices to minimize an objective function $\mathcal{L}$ with three components: 
\begin{equation}
    \mathcal{L} = \mathcal{L}_\text{n} + \mathcal{L}_{\nabla n} + \mathcal{L}_\text{sil}.
\end{equation} 
For simplicity, we will slightly abuse the notation and define $\norm{a - b}_1$ as the sum of the absolute differences
between all the pixels in image $a$ and image $b$. Thus,
$\mathcal{L}_\text{n}$ is the $\ell_1$ loss between the normal rendering of the mesh and the normal
generated by $\mathcal{G}^{\text{mv}}$: 
\begin{equation}
    \mathcal{L}_\text{n} = \frac{1}{|\Theta|} \sum_{\theta \in \Theta}\norm{\mathcal{R}_n[M,\theta] - I_{n,\theta}}_1.
\end{equation}  
$\mathcal{L}_{\nabla n}$ is the $\ell_1$ loss between the image gradient of the normal rendering of the mesh and the image gradient of the target normal:
\begin{equation}
    \mathcal{L}_{\nabla n} = \frac{1}{|\Theta|} \sum_{\theta \in \Theta}
    \norm{\frac{\partial \mathcal{R}_n[M,\theta]}{\partial x} - \frac{\partial I_{n,\theta}}{\partial x}}_1 +
    \norm{\frac{\partial \mathcal{R}_n[M,\theta]}{\partial y} - \frac{\partial I_{n,\theta}}{\partial y}}_1.
\end{equation}

\begin{figure}[t]
  \centering
  \includegraphics[width=\linewidth]{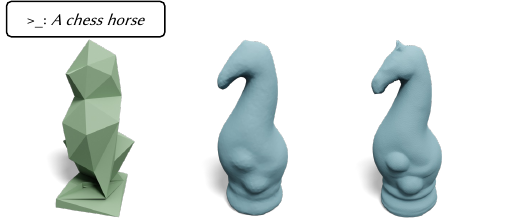} \\
  \vspace{-5.0mm}
  \begin{subfigure}[t]{0.325\linewidth}
    \caption*{Input 3D mesh}
  \end{subfigure}
  \hfill
  \begin{subfigure}[t]{0.325\linewidth}
    \caption*{Ours w/o $\mathcal{L}_{\nabla n}$}
  \end{subfigure}
  \hfill
  \begin{subfigure}[t]{0.325\linewidth}
    \caption*{Ours}
  \end{subfigure}
  \\
  \vspace{-3.0mm}
  \caption{
  \textbf{Ablating $\mathcal{L}_{\nabla n}$}.
  Adding the $\mathcal{L}_{\nabla n}$ component during mesh optimization allows the final mesh to
  incorporate more details from the generated normal images.
  In this example, notice the ears of the horse and the crisper details on the bottom of the piece.
  }
  \label{fig:loss_ablation}
\end{figure}

We found experimentally that $\mathcal{L}_{\nabla n}$ is a crucial component for capturing the higher frequency details
present in the generated normal images.
See Figure~\ref{fig:loss_ablation} for an example comparing the mesh optimization procedure with and without this
component.
$\mathcal{L}_\text{sil}$ is the $\ell_1$ loss between the silhouette rendering $\mathcal{R}_s$ of the mesh and the foreground segmentation mask $\mathcal{S}[I_{\text{rgb},\theta}]$ predicted from the RGB image $I_{\text{rgb},\theta}$ (generated by $\mathcal{G}^{\text{mv}}$) using Segment Anything~\citep{SAM}:
\begin{equation}
    \mathcal{L}_\text{sil} = \frac{1}{|\Theta|} \sum_{\theta \in \Theta}\norm{\mathcal{R}_s[M,\theta] - \mathcal{S}[I_{\text{rgb},\theta}]}_1.
\end{equation}

After each optimization iteration, we follow continuous remeshing~\citep{continuous_remeshing} to adaptively remesh the mesh $M$.

\paragraph{Implementation details.} 
We implement our mesh optimization algorithm in PyTorch. 
We run mesh optimization for 2,000 iterations. 
The mesh optimization process takes around 20 seconds on one single NVIDIA 3080Ti GPU. 

\section{Experiments}

In this section, we analyze the capabilities of our method and investigate how modifying various components
impacts its overall performance.
We start by presenting a quantitative and qualitative comparison with the current state-of-the-art in the
text-guided mesh refinement task.
Then, we show how the user can apply different guidance levels to trade off between detail generation and original
shape preservation.
Finally, we present a few additional results and applications.
We will make our code and trained model weights available to the public upon acceptance. 

\bgroup
\def\arraystretch{1.3}
\begin{table}[t]
  \begin{center}
    \caption{
    \textbf{Quantitative results.}
    Our method achieves the best results on CLIP similarity while being approximately two orders of magnitude
    faster.
    }
    \label{tab:quantitative}
    \vspace{-3.0mm}
    \begin{tabular}{lrr}
      \toprule
      \rowcolor{LavenderBlue}
      Method & CLIP similarity $\uparrow$ & Run time (sec) $\downarrow$ \\
      \midrule 
      Magic3D     & 0.2327 $\pm$ 0.0393 & 5597 \\
      \rowcolor[HTML]{EFEFEF}
      Latent-NeRF & 0.2367 $\pm$ 0.0415 & 3697 \\
      Fantasia3D  & 0.2556 $\pm$ 0.0335 & 2971 \\
      \rowcolor[HTML]{EFEFEF}
      Ours        & \textbf{0.2666 $\pm$ 0.0317} & \textbf{32} \\
      \bottomrule
    \end{tabular}
  \end{center}
\end{table}
\egroup

\subsection{Text-guided Mesh Refinement}

\paragraph{Baseline methods.} 

While there are methods that rely on supervisory signals from joint text-image embeddings~\citep{TextDeformer,CLIP_Mesh}, recent work has shown that better reconstruction quality comes from methods using score distillation sampling (SDS)~\citep{DreamFusion}. 
For this reason, we mainly compare with methods using SDS. 
We choose Latent-NeRF~\citep{LatentNeRF}, Magic3D~\citep{Magic3D} and Fantasia3D~\citep{Fantasia3D} as the competing methods. 
Latent-NeRF introduces global shape guidance through a sketch shape and a loss between the underlying NeRF's occupancy value and the winding number to the sketch shape's surface. We use their official implementation and their default setting to generate results for comparison with our method. 
Fantasia3D has their shape-guided 3D generation procedure implemented by initializing their initial SDF representation with an input shape.
However, they don't enforce any control mechanism during the generation. 
As a result, their generation can diverge from the input shape's outline. 
Magic3D introduced a two-stage method for text-to-3D generation. 
Their second stage refines the coarse shape generated from the first stage via direct mesh optimization with differentiable rendering and it is therefore comparable to our method.
For a fair comparison, we remove the first stage and initialize the Magic3D's second stage geometry with the guidance shape in a randomly initialized color field. 
For both Magic3D and Fantasia3D, we use the public implementation by threestudio~\citep{threestudio2023}.
Notice that all these methods rely on the score distillation sampling which usually takes more than 30 minutes per generation while our method takes less than a minute. 
For a runtime comparison, please refer to Table~\ref{tab:quantitative}.

\paragraph{Quantitative results.} 
Table~\ref{tab:quantitative} reports the CLIP~\citep{CLIP} similarity result and the run time of each method. 
We use the public implementation of the CLIP similarity metric from \citet{CLIPScore}. 
This metric computes the cosine similarity between the text prompt and images of the final meshes. 
Our method achieves the best CLIP similarity while running almost two orders of magnitude faster.

\begin{figure*}[t]
  \centering
  \vspace{-5.0mm}
  \includegraphics[width=0.98\linewidth]{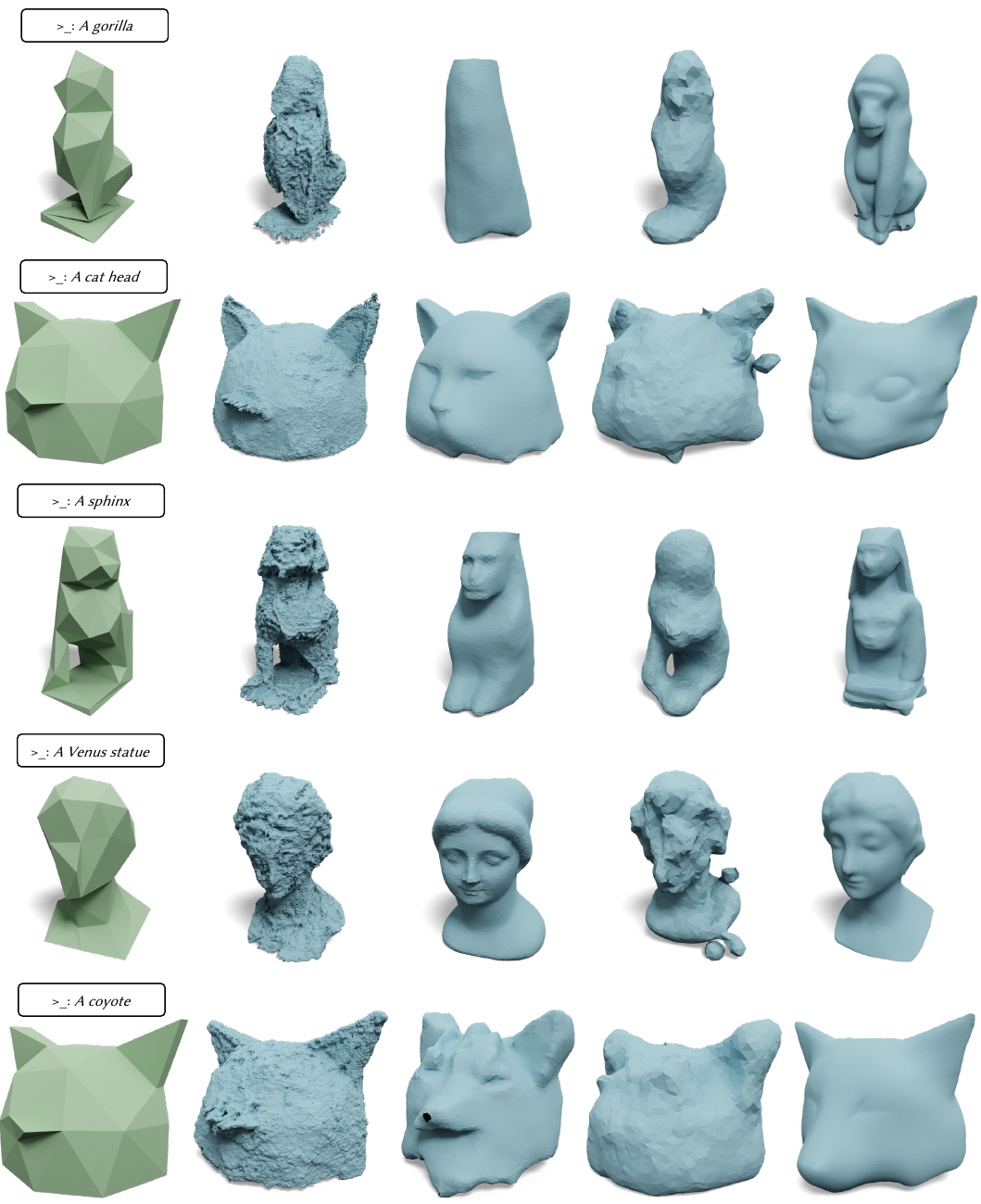} \\
  \vspace{-5.0mm}
  \begin{subfigure}[t]{0.196\linewidth}
    \caption*{Input 3D mesh}
  \end{subfigure}
  \hfill
  \begin{subfigure}[t]{0.196\linewidth}
    \caption*{Latent-NeRF}
  \end{subfigure}
  \hfill
  \begin{subfigure}[t]{0.196\linewidth}
    \caption*{Fantasia3D}
  \end{subfigure}
  \hfill
  \begin{subfigure}[t]{0.196\linewidth}
    \caption*{Magic3D}
  \end{subfigure}
  \hfill
  \begin{subfigure}[t]{0.196\linewidth}
    \caption*{Ours}
  \end{subfigure}
  \\
  \vspace{-3.0mm}
  \caption{
  \textbf{Qualitative results.} 
  Our method generates 3D meshes that have better geometric details and visual quality compared to state-of-the-art methods.
  }
  \label{app-fig:experiment-1}
\end{figure*}

\paragraph{Perceptual study and qualitative results.} 
We also present a casual perceptual study comparing our method with the three competing baselines. 
We designed 18 questions applied to 41 different participants. 
For each question, we present to the participant an input coarse mesh, a text prompt, a mesh refined by our method and a mesh refined by one of the three baseline methods. 
We ask the participant to select the mesh that has better geometric details, has better visual quality, and is more consistent with the input text prompt. 
Based on the results, we compute the preference, i.e., the percentage of participants who prefer the mesh refined by our method. 
Table~\ref{tab:user_study} reports the study results. 
Figure~\ref{fig:experiment} shows the qualitative results of each method. 
More visual results are provided in Figure~\ref{app-fig:experiment-1} and Figure~\ref{app-fig:experiment-2}. 
The perceptual study result indicates that our method is favored across all dimensions analyzed. 
Qualitatively, Latent-NeRF uses a NeRF representation that notoriously has problems generating surfaces since it is fundamentally a
volumetric occupancy representation. 
While Magic3D also suffers from generating meaningful results. 
Fantasia3D is based on SDS and generates visually good results at the expense of a long run time. 
Our method generates results that are more consistent with the input 3D mesh structure than the other methods and is much faster since we only need to run inference on pre-trained networks. 
The mesh refinement step operates directly on the input mesh and converges within seconds. 

\bgroup
\def\arraystretch{1.3}
\begin{table}[t]
  \begin{center}
    \caption{
    \textbf{User study results.}
    Our method was selected as the favorite for the majority of users in all evaluated setups.
    }
    \vspace{-3.0mm}
    \label{tab:user_study}
    \begin{tabular}{lr}
      \toprule
      \rowcolor{LavenderBlue}
      Comparison & Preference \\
      \midrule 
      Ours vs. Magic3D     & \\
      \rowcolor[HTML]{EFEFEF}
      $\;\;\;\bullet\;$ Better geometric details & 96.35\% \\
      $\;\;\;\bullet\;$ Better visual quality & 100.00\% \\
      \rowcolor[HTML]{EFEFEF}
      $\;\;\;\bullet\;$ More consistent with input text & 98.80\% \\
      Ours vs. Latent-NeRF & \\
      \rowcolor[HTML]{EFEFEF}
      $\;\;\;\bullet\;$ Better geometric details & 89.05\% \\
      $\;\;\;\bullet\;$ Better visual quality & 93.90\% \\
      \rowcolor[HTML]{EFEFEF}
      $\;\;\;\bullet\;$ More consistent with input text & 76.80\% \\
      Ours vs. Fantasia3D  & \\
      \rowcolor[HTML]{EFEFEF}
      $\;\;\;\bullet\;$ Better geometric details & 84.15\% \\
      $\;\;\;\bullet\;$ Better visual quality & 65.85\% \\
      \rowcolor[HTML]{EFEFEF}
      $\;\;\;\bullet\;$ More consistent with input text & 69.50\% \\
      \bottomrule
    \end{tabular}
  \end{center}
\end{table}
\egroup

\subsection{Controllability}
\label{sec:controlability}

\paragraph{Control guidance strength.}
We allow the users to control how much detail is generated by allowing them to select how many percentage of the backward diffusion process will use the given control guidance. 
As seen in Figure~\ref{fig:guidance_strength}, this can happen in both single-view and multi-view stages.
During the single-view stage, not using $\mathcal{R}_d[M_\text{coarse}, \theta_s]$ means creating an image
uncorrelated to $M_\text{coarse}$, not desirable for our application.
Using the control guidance throughout the whole generative process, on the other hand, may result in an image that
follows $M_\text{coarse}$ too strictly which might incur in artificial shapes like the squirrel to the right in Figure~\ref{fig:guidance_strength}.
During the multi-view stage, the guidance will control how many percentage of the multi-view generation process uses images $\{\mathcal{R}_n[M_\text{coarse}, \theta]\}_{\theta \in \Theta}$ as control guidance -- when not used at all (0\% in the bottom row of Figure~\ref{fig:guidance_strength}), our multi-view generation
stage is reduced to Wonder3D.
As the control guidance increases, information from other views becomes more relevant, but using control guidance 100\% of the denoising steps
(100\% in the bottom row of Figure~\ref{fig:guidance_strength}) yields a final shape that is very similar to $M_\text{coarse}$
without too many added details.
We empirically found that values between 30\% and 75\% strike a good balance between controllability and detail generation.

\paragraph{Pose control.}
Our method allows users to provide explicit control over the pose of the resulting 3D mesh.
Figure~\ref{app-fig:pose_control} shows three examples, each of which has a different pose. 
Our method generates 3D meshes that have more geometric details and are consistent with the pose of the input 3D mesh.

\subsection{Application}

We develop a demo that builds on the Gradio library to provide users with a user-friendly interface for 3D modeling applications. 
See the supplemental video for the demo.  
We will release our demo upon paper acceptance.

\begin{figure}[t]
  \centering
  \includegraphics[width=\linewidth]{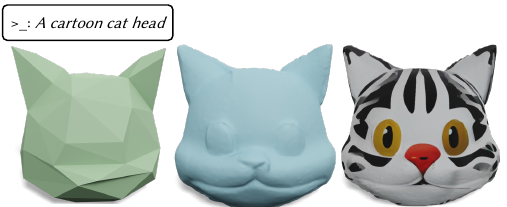} \\
  \vspace{-5.0mm}
  \begin{subfigure}[t]{0.325\linewidth}
    \caption*{Input 3D mesh}
  \end{subfigure}
  \hfill
  \begin{subfigure}[t]{0.325\linewidth}
    \caption*{Ours}
  \end{subfigure}
  \hfill
  \begin{subfigure}[t]{0.325\linewidth}
    \caption*{Ours with texture}
  \end{subfigure}
  \\
  \vspace{-3.0mm}
  \caption{
  \textbf{Mesh texturing.} 
  Our final refined mesh can also be textured.
  }
  \label{fig:texturing}
\end{figure}

\subsection{Mesh Texturing}

While the goal of our method is to add geometric details to input 3D meshes, our method can also be used to texture a mesh. 
Since our method builds on Wonder3D which outputs both multi-view RGB images and normals, we can use the multi-view RGB predictions to texture the output 3D mesh. 
We use the mesh texturing code from Direct2.5~\citep{direct2.5}. 
We refer readers to \citet{direct2.5} for details about mesh texturing.
Figure~\ref{fig:texturing} shows an example. 

\section{Conclusions}
In this paper, we offer a novel fast method for adding high-quality geometric details to coarse 3D meshes. 
Our key design decision is to rely only on feed-forward networks, generating a single-view RGB image conditioned on the coarse input mesh, to further use as a condition for multi-view ControlNet that we train to generate multi-view consistent normal images. 
Finally, we can use the resulting normal images to refine the input coarse geometry in a matter of seconds, since we only need differentiable rasterization, and do not rely on propagating gradients via any diffusion networks. 
We demonstrate that our method is useful in several scenarios, including refining low-poly shapes and shapes produced from an arrangement of simple primitives, and also supports texturing the refined mesh.
In addition to speed and quality, our method offers explicit control over the pose and the coarse structure of the output, which we believe is essential for real creative applications. 

\begin{figure*}[t]
  \centering
  \includegraphics[width=\linewidth]{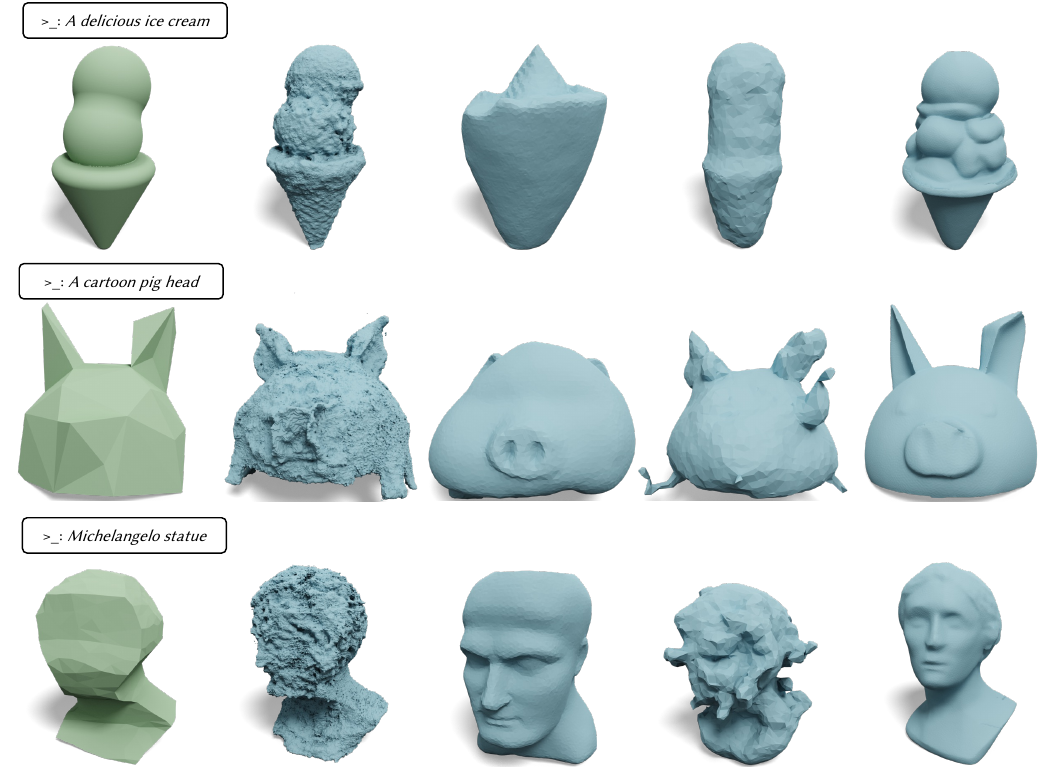} \\
  \vspace{-5.0mm}
  \begin{subfigure}[t]{0.196\linewidth}
    \caption*{Input 3D mesh}
  \end{subfigure}
  \hfill
  \begin{subfigure}[t]{0.196\linewidth}
    \caption*{Latent-NeRF}
  \end{subfigure}
  \hfill
  \begin{subfigure}[t]{0.196\linewidth}
    \caption*{Fantasia3D}
  \end{subfigure}
  \hfill
  \begin{subfigure}[t]{0.196\linewidth}
    \caption*{Magic3D}
  \end{subfigure}
  \hfill
  \begin{subfigure}[t]{0.196\linewidth}
    \caption*{Ours}
  \end{subfigure}
  \\
  \vspace{-3.0mm}
  \caption{
  \textbf{Qualitative results.} 
  Our method generates 3D meshes that have better geometric details and visual quality compared to state-of-the-art methods.
  }
  \label{app-fig:experiment-2}
  \vspace{2.0mm}
  \begin{minipage}[t]{0.49\linewidth}
    \includegraphics[width=\linewidth]{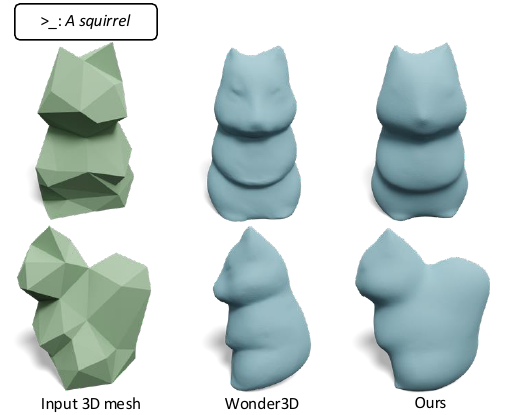}
    \vspace{-7.0mm}
    \caption{
    \textbf{Multi-view control.}
    Our method is capable of generating details using the full initial shape as guidance. 
    On the other hand, Wonder3D results yield reasonable renditions when visualized from the initial viewpoint $\theta_s$ (top row) but clearly fail to follow the coarse geometric guidance when seen from other views (bottom row).
    }
    \label{app-fig:mv_control}
  \end{minipage}
  \hfill 
  \begin{minipage}[t]{0.49\linewidth}
    \includegraphics[width=\linewidth]{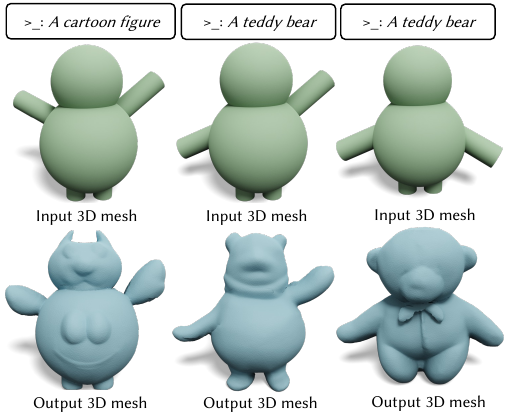}
    \vspace{-7.0mm}
    \caption{
    \textbf{Pose control.}
    Our method is capable of generating meshes that follow the pose of the input mesh.
    }
    \label{app-fig:pose_control}
  \end{minipage}
\end{figure*}

\paragraph{Limitations and future work.}
Our method was trained to generate 6 images of 256 $\times$ 256 resolution for the subsequent mesh refinement. 
The number of views and the image resolution limit the level of geometric details our method can generate. 
In addition, our mesh refinement algorithm relies on the segmentation quality of an off-the-shelf image segmentation model.
The refinement produces artifacts when the segmentation is faulty or inconsistent across different views. 
Training a model to generate more view-consistent images with higher resolution, as well as employing more robust image segmentation models, can boost the performance of our method. 
There are other interesting topics raised during the development of our method, such as image-guided mesh refinement, localized detail generation in selected regions or parts, and identity preservation when changing the pose of the coarse mesh (see Figure~\ref{app-fig:pose_control}), which we aim to explore in the future.

\begin{acks}
This paper is funded in part by NSERC Discovery (RGPIN–2022–04680), the Ontario Early Research Award program, the Canada Research Chairs Program, a Sloan Research Fellowship, the DSI Catalyst Grant program and gifts by Adobe Inc.

We thank Chenxi Liu, Otman Benchekroun, Joonho Kim, Kinjal Parikh, Kevin Mu, Abhishek Madan, Gabrielle Browne, Wenzhi Guo, Aravind Ramakrishnan, Victor Rong, Yixin Chen, Lily Goli, Yuta Noma, Anagh Malik, Chang Yue, Chao-Te Chou, Dylan Turpin, Tsun-Yi Yang, Yuliang Zou, Chen Gao, Shih-Yang Su and Wendy Choi for helping with the perceptual study, and Sven Dickinson, Sanja Fidler, Ziyi Wu, Shih-Yang Su and Yuliang Zou for insightful conversations that inspired us in this work. 
We would like to acknowledge the Vector Institute for the computing support. 
We thank Xuan Dam, John Hancock and all the University of Toronto Department of Computer Science research, administrative and maintenance staff. 
\end{acks}

\bibliographystyle{ACM-Reference-Format}
\bibliography{references}

\end{document}